\documentclass[letterpaper, 10 pt, conference]{ieeeconf}  % Comment this line out if you need a4paper

\IEEEoverridecommandlockouts                              % This command is only needed if 
\usepackage{amsmath,amsfonts}
\usepackage{algorithmic}
\usepackage{algorithm}
\usepackage{array}
\usepackage[caption=false,font=normalsize,labelfont=sf,textfont=sf]{subfig}
\usepackage{textcomp}
\usepackage{stfloats}
\usepackage{url}
\usepackage{verbatim}
\usepackage{graphicx}
\usepackage{cite}

\begin{document}
\title{NDD: A 3D Point Cloud Descriptor Based on Normal Distribution for Loop Closure Detection}

% \author{BIRL \\
%         Rui-hao Zhou, Xu-bin Lin, Hong Zhang, Yi-sheng Guan, and Li He*}
       
        % <-this % stops a space

\author{Ruihao Zhou$^{1}$, Li He$^{2}$, Hong Zhang$^{2}$, Xubin Lin$^{1}$, Yisheng Guan$^{1}$*
% <-this % stops a space
\thanks{This work was supported in part by the National Natural Science Foundation of China under Grant No. 62173096, in part by the Leading Talents Program of Guangdong Province under Grant No. 2016LJ06G498 and 2019QN01X761, in part by Guangdong Yangfan Program for Innovative and Entrepreneurial Teams under Grant No. 2017YT05G026.} %This work was supported by National Natural Science Foundation of China (NSFC) Grant No. 6217020307, and Guangdong Province Special Fund for Modern Agricultural Industry Common Key Technology R\&D Innovation Team under Grant No. 2019KJ129 .}<-this % stops a space
\thanks{$^{1}$Ruihao Zhou, Xubin Lin and Yisheng Guan is with the Department of Electromechanical Engineering, 
Guangdong University of technology, China. Corresponding author: Yisheng Guan (ysguan@gdut.edu.cn).}
        % {\tt\small albert.author@papercept.net}}%
\thanks{$^{2}$Li He and Hong Zhang (\{hel, hzhang\}@sustech.edu.cn) are with the Department of Electronic and Electrical Engineering, Southern University of Science and Technology, China, and with Shenzhen Key Laboratory of Robotics and Computer Vision.}
%\thanks{*https://github.com/zhouruihao1001/NDD}
        % Dayton, OH 45435, USA
        % {\tt\small b.d.researcher@ieee.org}}%
}
%\thanks{This paper was produced by the IEEE Publication Technology Group. They are in Piscataway, NJ.}% <-this % stops a space
%\thanks{Manuscript received April 19, 2021; revised August 16, 2021.}}

% \markboth{Journal of \LaTeX\ Class Files,~Vol.~14, No.~8, March~2022}%


\maketitle

\begin{abstract}
Loop closure detection is a key technology for long-term robot navigation in complex environments. In this paper, we present a global descriptor, named Normal Distribution Descriptor (NDD), for 3D point cloud loop closure detection. The descriptor encodes both the probability density score and entropy of a point cloud as the descriptor. We also propose a fast rotation alignment process and use correlation coefficient as the similarity between descriptors. Experimental results show that our approach outperforms the state-of-the-art point cloud descriptors in both accuracy and efficency. The source code is available and can be integrated into existing LiDAR odometry and mapping (LOAM) systems.

\end{abstract}

% \begin{IEEEkeywords}
% point cloud, loop closure detection, range sensing, SLAM.
%Article submission, IEEE, IEEEtran, journal, \LaTeX, paper, template, typesetting.
% \end{IEEEkeywords}

\section{Introduction}
% \IEEEPARstart{S}{imultaneous} 
Simultaneous localization and mapping (SLAM) is one of the basic technologies in robotics that has developed rapidly in recent decades. Compared with visual SLAM, LiDAR SLAM is robust to illumination changes and hence is widely used in localization tasks. It is important for loop closure detection in SLAM, i.e., whether a robot has returned to a previously visited scene. Point cloud loop closure detection always needs to describe one point cloud of a scene in terms of a descriptor. When a robot returns to a location it has visited, the loop closure is determined by similarity calculation between the current descriptor and candidate descriptors.

Typically, there exist three classes of descriptors to represent a point cloud: bag-of-words, hand-craft global descriptor and and deep-learning-based global descriptor. Bag-of-words (BoW) \cite{ref_bow} was adopted in LIDAR settings due to its success in image retrieval. BoW ﬁrst calculates local descriptors of key points, and vector-quantizes the local descriptors into words of a dictionary that has been constructed off line. The descriptor uses the histogram of the words, which has high complexity due to vector quantization or clustering. In addition to bag-of-words, there are several hand-crafted descriptors such as Viewpoint Feature Histogram (VFH) \cite{ref_vfh}, Small-Sized Signatures (Z-projection) \cite{ref_zprojection} and Ensemble of Shape Functions (ESF) \cite{ref_esf}. Compared with the BoW method and its variants, hand-crafted global descriptors avoid the time-consuming local key point detection part. However, global descriptors are sometimes sensitive to laser sensor orientation and position changes. This is due to the lack of global structure and local distribution in building such descriptors.

In recent years, some works focus on describing point clouds by learning. Learning-based descriptors show great performance on precision, but require high computational cost and memory with GPU. These methods are usually difficult to apply to unseen environments due to poor generalization.

In this paper, we propose a method, named normal distribution descriptor (NDD), to describe a 3D point cloud as a global descriptor for LiDAR-based loop closure detection. We represent one point cloud by both its probability density score and entropy. In addition, we propose a fast retrieval process and replace the popular cosine similarity with the correlation coefficient similarity. The framework of our method is shown in Fig. 1. The main contributions of this work are:
% \subsubsection*{\bf A simple numbered list}
\begin{itemize}
% [leftmargin=*]
\item{Using two-scale statistics based on normal distribution as features to describe a point cloud.}
\item{Replacing cosine distance with correlation coefficient and proposing a fast alignment method between two descriptors for similarity matching.}

\end{itemize}

\section{Related Work }
Point cloud descriptors are generally divided into two classes: local and global. Common local point cloud description algorithms include normal line and curvature calculation, eigenvalue analysis, PFH\cite{ref_pfh}, Spin Image\cite{ref_spin}, shape contexts\cite{ref_shape contexts}, SHOT\cite{ref_shot}, etc. SHOT constructs a spherical region with feature points as the center and divides this sphere into cells. Histogram statistics of points in each cell are used as the descriptor. One can convert a local descriptor into a global one by expanding the support space for a point cloud.

Scan Context (SC)\cite{ref_sc} is one of the most popular hand-crafted global descriptors. SC is a signature-based global descriptor that directly records the vertical information of the 3D space and does not require prior knowledge. In SC, one point cloud is encoded by the distribution of cell maximum heights. Using the height rather than the shape of the structure eliminates variations in sparsity effected by sensing resolution, distance, and different scales. Variants of Scan Context have been proposed in recent years. \cite{ref_isc}\cite{ref_vo} show the feasibility by using the intensity or relative distance of 3D points. 
Wang\cite{ref_iris} shows that a binary signature image can be obtained for each point cloud after several LoG-Gabor ﬁltering and thresholding operations on the LiDAR-Iris image representation. M2DP\cite{ref_m2dp} projects one point cloud onto several specific two-dimensional planes and generates histograms, which are combined to form a matrix. Then Singular Value Decomposition (SVD) is applied to this matrix, and the first left and right singular vectors are concatenated as the final descriptor. This method does not need to calculate point cloud normals and has high efficiency and robustness.

Learning is a popular way in generating a global descriptor. PointNetVLAD\cite{ref_pv} extracts global descriptors from a given 3D point cloud through end-to-end training and shows high accuracy and robustness. LPD-Net\cite{ref_lpd} adaptively selects $K$ values to extract a global descriptor with recognition and generalization when constructing a neighborhood map using $K$ Nearest Neighbors (KNN). SeqLPD\cite{ref_seq} is a neural network to extract global descriptors, where sequence matching reduces the computational time by selecting several super keyframes. GOSMatch\cite{ref_gos} is generated from the spatial relationship between semantics, and is applied to frame description and data association. However, a limitation of the learning-based approach is that they sometimes show poor generalization\cite{ref_gen}. In addition, these learning-based methods need a lot of training data. So, in this paper, we focus on a hand-craft global descriptor.

\section{Descriptor Generation}

% \textbf{Overview.}
In this section, we present a novel 3D point cloud global descriptor in bird’s eye view (BEV), named normal distribution descriptor (NDD).

\begin{figure*}[!t]
\centering
\includegraphics[width=6.5in]{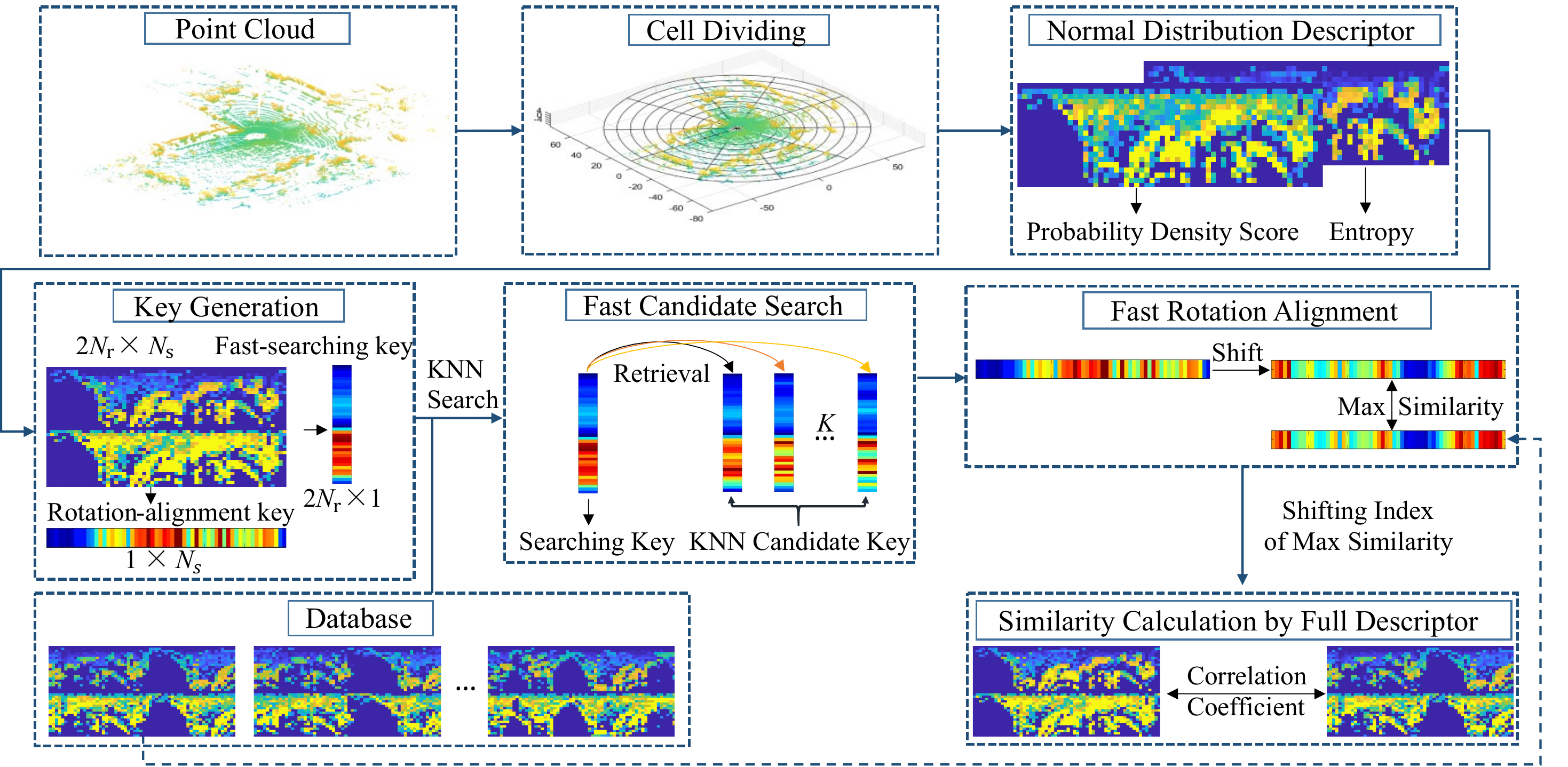}
\caption{Approach overview. First, one point cloud is divided into cells and one cell is represented by its probability density score and entropy. Then, A vector named searching key is compressed from full descriptor and used to create a KD-tree for retrieving the nearest neighbors. Finally, the retrieved nearest neighbor descriptors in the database are matched with the query descriptor, in the form of full descriptor, after fast rotation alignment.}
% It is considered a loop when the most similar candidate descriptor with the query satisﬁes the threshold.
\label{fig_overview}
\end{figure*}

\subsection{Preprocessing}
Firstly, we uniformly sample of a raw point cloud for reducing computing cost. 
% The type of each point cloud is determined according to the situation of the eigenvalue size\cite{ref_prepocess}. 
We assume that each point cloud has a dominant direction\cite{ref_prepocess}. Therefore, we use PCA  (Principal Component Analysis) to align a point cloud to roughly realize rotation invariance. In this case, the first PC is considered as the x-axis and the second as y-axis.

\subsection{Cell Dividing and Encoding}
% \textbf{Cell Dividing and Encoding.} 
Inspired by\cite{ref_shape contexts}\cite{ref_sc}, we divide equally a point cloud into many cells in polar coordinate, as shown in Fig. 2. One scan can be converted into $N_{r}$ rings and $N_{s}$ sectors in radial and azimuthal directions, respectively. 
%So, each cell is represented by a cell signature: $D_{i j}, i \in N_{r}, j \in N_{s}$.
% For each cell of measurement $D_{i j}$, we assign values using an encoding function $\psi(D_{i j})$.
\begin{figure}[!t]
\centering
\includegraphics[width=2.8in]{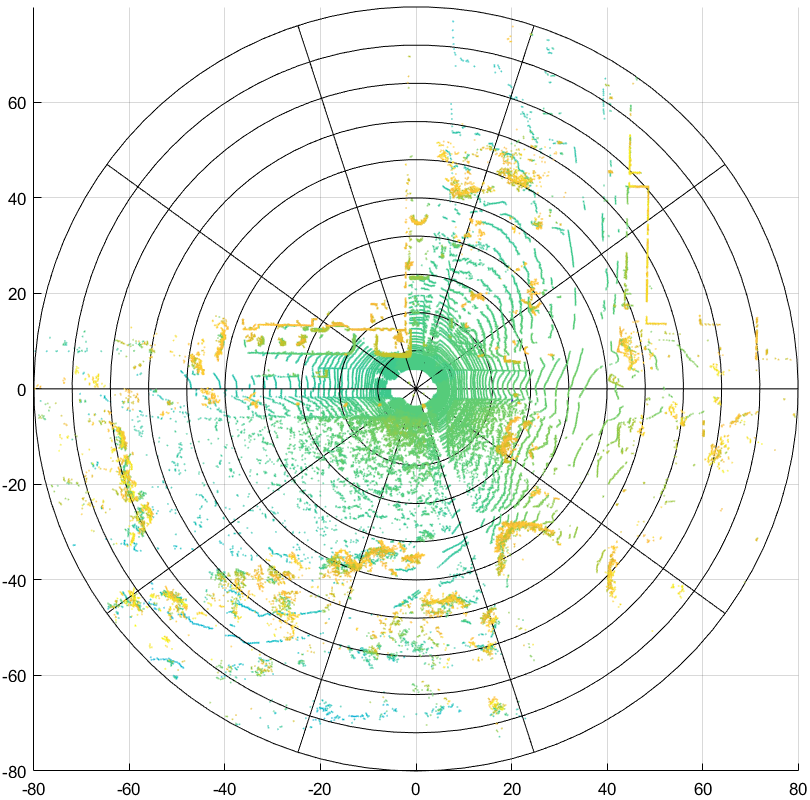}
\caption{Cell division of a BEV representation}
\label{fig_2}
\end{figure}

% According to previous studies, most descriptors will highly generalize the local feature of point clouds, resulting in a great loss of local structural information. 
Many previous descriptors use simple statistics in one cell, such as maximum height\cite{ref_sc}, intensity\cite{ref_isc} and the number of points\cite{ref_m2dp}. However, describing a point cloud by local extrema sometimes does not show good robustness to external environmental changes and noise.

To describe the distribution of points in a cell, we employ the normal distribution, $C\sim(\mu, \Sigma)$, where $\mu$ is the mean and $\Sigma$ is the covariance of points in a cell, to encode the local shape\cite{ref_ndt}. In our approach, we describe all the points in one cell with two statistics of a normal distribution. The first is probability density score:

\begin{equation}
\label{deqn_ex_P}
P=\sum_{i}^{N} \exp \left(-\frac{\left(x_{i}-\mu\right)^{\prime} \Sigma^{-1}\left(x_{i}-\mu\right)}{2}\right),
\end{equation}
where $N$ is the number of points in one cell, $x_i$ is the 3D data of a point. The second is entropy:
\begin{equation}
\label{deqn_ex_H}
E=-\int p(x) \log p(x) d x=\frac{d}{2}(\log 2 \pi+1)+\frac{1}{2} \log |\Sigma|,
\end{equation}
where $|\Sigma|$ denotes the matrix determinant and $d$ is the dimension of the vector space. In this paper, $d=3$ for 3D point cloud.

% Finally, we describe a single 3D point cloud with two dimension matrices $(N_{r} \times N_{s} \times 2)$ as illustrated in Fig. 3. 
Let $P_c$ denote the $N_{r} \times N_{s}$ matrix containing $P$ of all cells, and $E_c$ of $E$, as shown in Fig. 3. So, the signature of the point cloud $D$ is:
\begin{figure}[!t]
\centering
\includegraphics[width=3in]{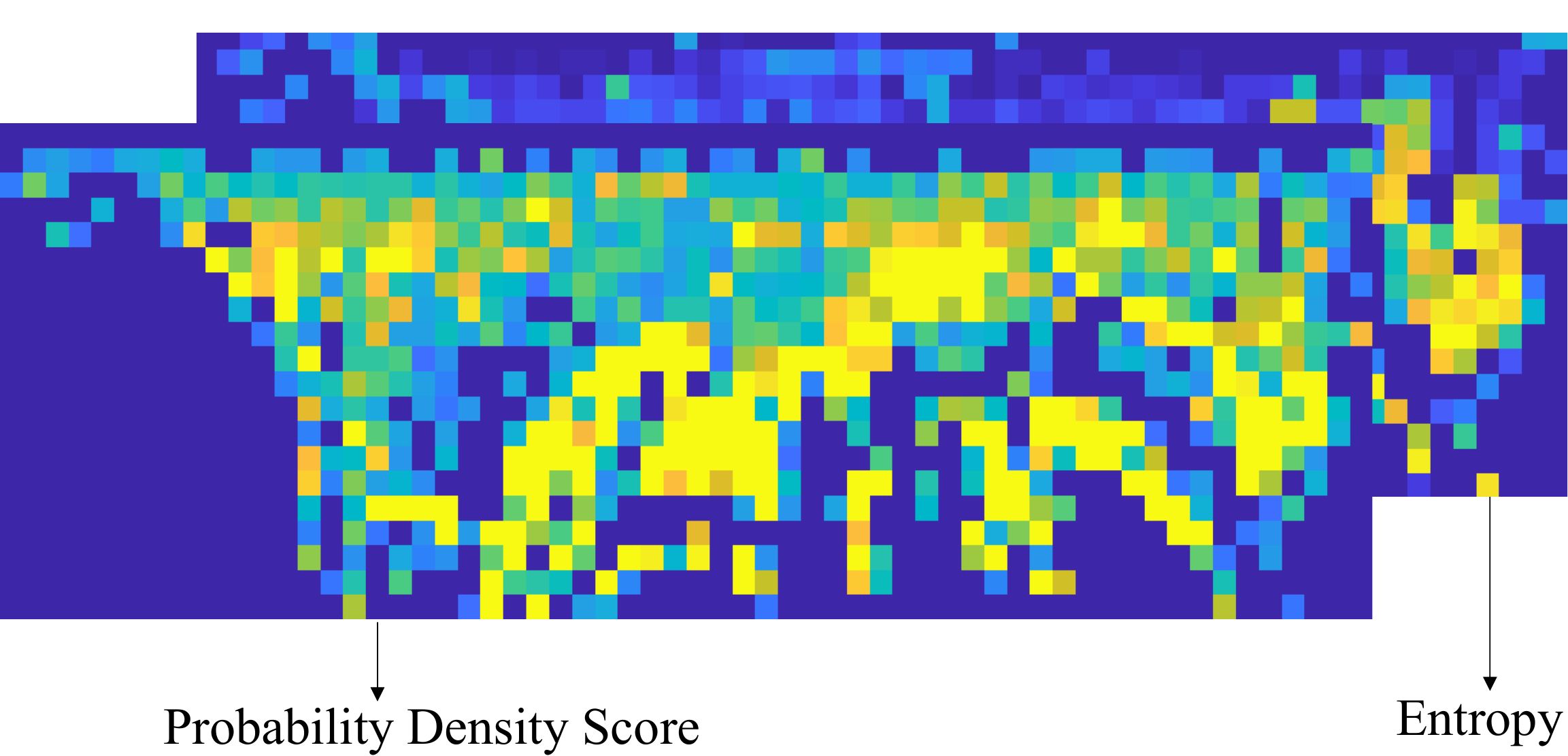}
\caption{Two-scale cell signature}
\label{fig_3}
\end{figure}

\begin{equation}
\label{deqn_ex_D}
\begin{aligned}
D=\left[\begin{array}{l}
                    P_{c} \\
                    E_{c}
                    \end{array}\right],
\end{aligned}
\end{equation}
which means we simply concatenate $P_c$ and $E_c$ to form the $2N_{r}\times N_{s}$ cell signature $D$. According to our experiments, the scale of probability density and that of entropy are similar. So, we adopt the simple concatenation fashion to combine both signatures.

Many previous works describe a point cloud by computing normals. It is usually hard to acquire an accurate normal from noisy 3D data. Using normal also needs high computational cost.

Based on the average statistics of the normal distribution, our approach reduces the effect of outliers in local data. 
In addition, probability density score can be considered as the sum of a natural exponent of Mahalanobis distance in local points.
Entropy, determined by covariance, describes the distribution of one cell. We summarize our descriptor, entitled Normal Distribution Descriptor (NDD), in Alg. 1.
% In addition, cell encoding by different scales can improve the performance of the descriptor.
\begin{algorithm}[htb] 
\caption{Normal Distribution Descriptor} 
\label{alg:Framwork} 
\begin{algorithmic}[1] 
\REQUIRE ~~ \\
Point cloud $P$, $N_s$ (number of the azimuth bins) and $N_r$ (number of the radial bins).\\
\ENSURE ~~\\ 
A $(2N_r)\times N_s$ matrix $D$.
\STATE Uniformly down-sampling over $P$;
\STATE Calculate the PCA of $P$. Align axis with the principal component;
\STATE Divide $P$ into $N_s$ azimuth and $N_r$ radial angle cells;
\FOR{$i = 1$ to $N_r$}
    \FOR{$j = 1$ to $N_s$}
    \STATE Calculate mean $\mu$ and covariance $\Sigma$ in the cell indexed by $i$ and $j$; 
    \STATE Calculate probability density score $(P_c)_{ij}$;
    \STATE Calculate entropy $(E_c)_{ij}$;
    %\STATE Assign $(P_c)_{ij} = P_{ij}, (E_c)_{ij} = E_{ij}$;
    \ENDFOR
\ENDFOR
    \STATE Assign $D=\left[\begin{array}{l}
                    P_c \\
                    E_c
                    \end{array}\right]$;
\RETURN $D$.
\end{algorithmic}
\end{algorithm}

\section{Retrieval and Matching}
In this section, we propose a fast retrieval method and use correlation coefficient to measure the similarity between two descriptors for loop closure detection.
\subsection{Key Generation and Fast Searching}
With the rapid growth of the map scale, the storage of point cloud descriptors also increases gradually. When a robot queries for closed loops, it requires high time cost to match the query descriptor with all saved descriptors in the database. Therefore, a KD-tree-based retrieval algorithm is used to achieve fast descriptor retrieval. The retrieval approach consists of two parts, key searching and KNN search.

\textbf{Key Searching.} It is computationally expensive to build and retrieve by the entire descriptors, with time complexity $O(nk\log{(n)})$ and $O(n^{(k-1)/k})$ respectively of building and searching in a KD-tree, where $n$ is the number of descriptors, and $k=2N_r N_s$ with the full descriptor in our works. So, we convert the signature $D$ into a $2N_r\times 1$ column vector $\gamma$, as the key of one descriptor, through summing over the rows of $D$ for fast searching.

We use $\gamma$ to build a KD-tree for fast key searching. As a result, construction complexity is reduced to $O(2N_r n\log n)$ from $O(2N_r N_s n\log n)$ of the full descriptor.
% \subsection{Two-Stage Loop Closure Detection}

\textbf{$K$ Nearest Neighbors Search.} Given a built key $\gamma$ KD-tree, we first search the query key in the tree and in return we obtain the $K$ nearest neighbors w.r.t their keys of the query. Then we compare the $K$ candidates with the query by using their full descriptors, in a purpose of accurate matching. There are two important issues in KNN search: rotation alignment and similarity calculation.

\subsection{Rotation Alignment}

Rotation alignment is an important task in matching two full descriptors. Like many other global descriptors, our NDD is rotation-sensitive. Intuitively, a rotated point cloud will result in a different column order in our descriptor. 

A popular rotation alignment technique is to fix the axes of both the reference and query point cloud by PCA. However, PCA cannot distinguish forward and reverse. In addition, PCA is not robust to large rotation transformation of a point cloud. In this paper, before matching, we use column-shifting to align two full descriptors.

\begin{figure}[!t]
\centering
\includegraphics[width=3.4in]{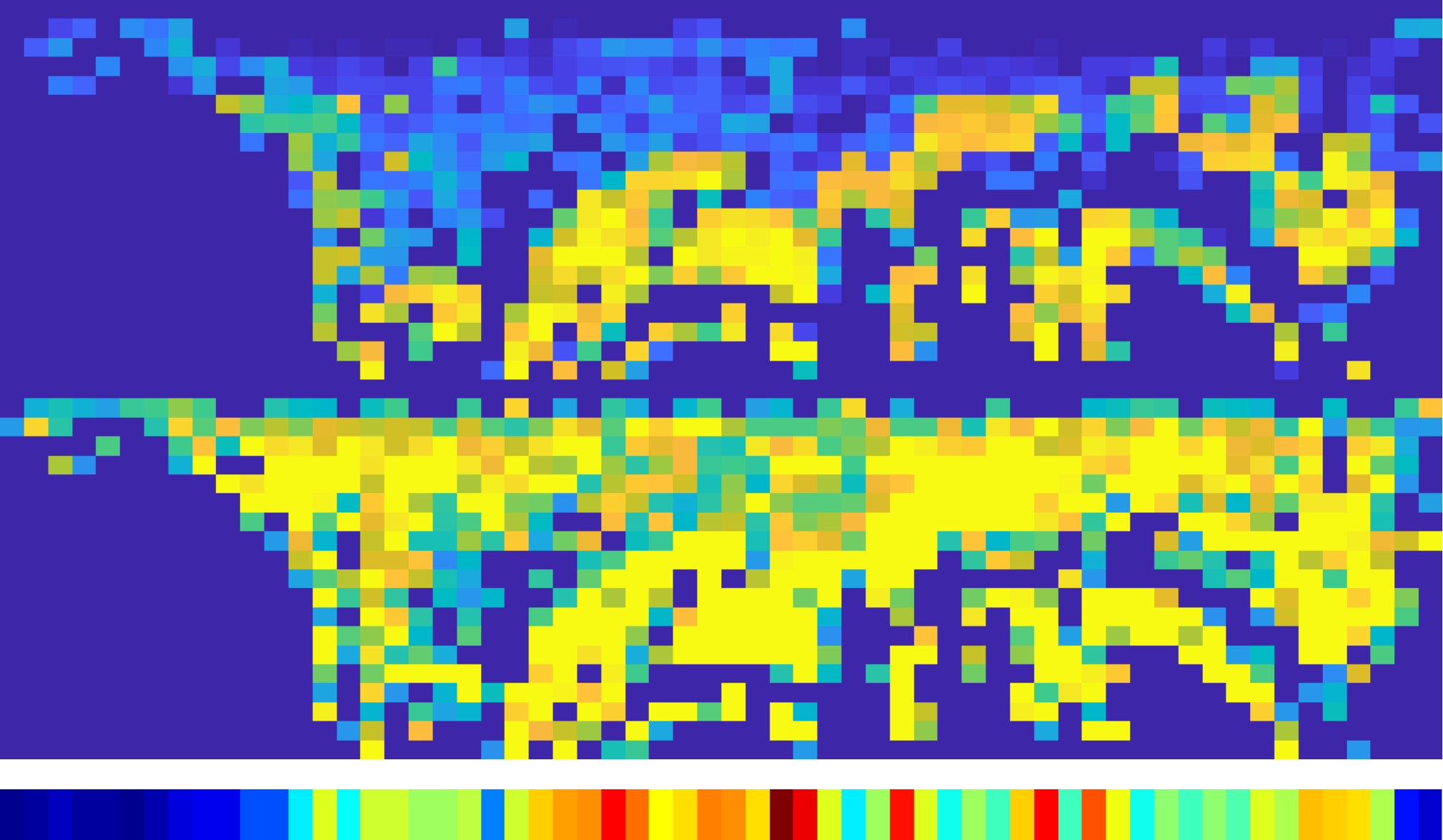}
\caption{Output descriptor $D$ and its column-wise sum $V$}
\label{fig_4}
\end{figure}

Previous column-shifting methods, such as the one used in Scan Context, directly shift two full descriptors. As a result, this method requires high computational cost. Instead of the full descriptor column-shifting, we sum up $D$ column-wise to form a $1\times N_s$ row vector $V$ as the rotation alignment key, as shown in Fig. 4.

% \begin{equation}
% \label{deqn_ex_V}
% V=\left(\xi_{1}, \xi_{2}, \cdots, \xi_{j}\right), j \in\left[1, N_{s}\right], j \in N^{*},
% \end{equation}
% where $\xi_{j}=\sum_{i=1}^{2 N_{r}} D_{i j}$.
Then we calculate the cosine similarity $S$ between two rotation alignment vectors $V_{A}$ and $V_{B}$ by:
\begin{equation}
\label{deqn_ex_O}
S\left(V_{A}, V_{B}\right)=\frac{V_{A} V_{B}^\mathrm{T}}{\left\|V_{A}\right\|\left\|V_{B}\right\|}.
\end{equation}

We calculate all possible $S(V_A, V_B)$ similarities between two rotation alignment vectors by column-shifting and take the index w.r.t. the maximum similarity as the result of rotation alignment. 

Compared with the full descriptor column-shifting, the computational complexity is reduced from $O(N_r N_s^2)$ to $O(N_s^2)$.

\subsection{Similarity Calculation Between NDDs}
% \textbf{Similarity Calculation Between NDDs. }
One-dimension matching, such as cosine or Euclidean distance, is the most popular fashion to estimate similarity between two descriptors. In our two-dimension descriptor, however, the performance of the one-dimension methods sometimes is not encouraging. Instead of cosine distance or other popular one-dimension matching methods, correlation coefficient between a query descriptor $D_A$ and the descriptor $D_B$ retrieved from the database is used to compute the similarity after rotation alignment. The correlation coefficient is:
\begin{equation}
\label{deqn_ex_corr}
r=\frac{\sum_{i j}\left[(D_A)_{ij}-\bar{a}\right]\left[(D_B)_{ij}-\bar{b}\right]}{\sqrt{\left\{\sum_{i j}\left[(D_A)_{ij}-\bar{a}\right]^{2}\right\}\left\{\sum_{i j}\left[(D_B)_{ij}-\bar{b}\right]^{2}\right\}}},
\end{equation}
where $\bar{a}$ is the mean of all elements in matrix $D_A$.

One drawback of the popular cosine similarity is that it only reflects difference of one dimensional vectors in direction. SC-fashion cosine distance converts a matrix into several column vectors and measures the average similarity of all columns in a matrix.
% which cause to the information loss of original matrix. So, SC-fashion cosine distance only measures the local differences of each column in matrix.
Correlation coefficient and cosine distance are identical when the origin is the arithmetic mean of the descriptors\cite{ref_corr1}\cite{ref_corr2}. 

Correlation coefficient is essentially a normalized measurement of the covariance. Unlike the cosine that is not invariant to shift, the correlation of two-dimensional descriptors is invariant to both scale and location changes. 

We summarize our retrieval and matching method in Alg. 2.

\begin{algorithm}[htb] 
\caption{NDD Matching} 
\label{alg:lcd} 
\begin{algorithmic}[1] 
\REQUIRE ~~ \\
Query point cloud descriptor $D$, and $K$ candidates from the KD-tree.\\
\ENSURE ~~\\ 
The max similarity of query descriptor and candidates.
\STATE Calculate row-wise sum of $D$ to form the query key $\gamma$;
\STATE Search the query key $\gamma$ in a pre-built KD-tree of the recorded key of point clouds; obtain the $K$ nearest neighbors;
% \STATE Search the tree and obtain the K nearest neighbors of the query; 
\STATE Calculate column-wise sum $V$ from the full descriptor $D$ and that of the $K$ nearest neighbors, and use the column-wise sum to find the index of column-shifting;
\STATE Calculate correlation coefficient $r$ between one of the $K$ candidates and the query with their full descriptor;

\RETURN Max similarity $r$ among all K candidates. 
\end{algorithmic}
\end{algorithm}

\section{Experiment}
% In this section, our representation and algorithm are evaluated over various datasets and against other representative algorithms. In addition, we integrate our NDD into slam system.
\subsection{Datasets and Experimental Settings}
\textbf{KITTI\cite{ref_kitti}}: The point clouds of KITTI dataset was scanned by a Velodyne HDL-64E LiDAR. Among the datasets, sequence 08 has reverse-only loops, sequence 02 has different direction loops, and sequence 00, 05, 06, 07 have same loops. We chose six sequences: 00, 02, 05, 06, 07, 08.

\textbf{Mulran\cite{ref_mulran}}: The Multimodal Range Dataset (MulRan) was captured by an Ouster 64-ray LiDAR. The dataset was designed for site identification tasks in urban cities regarding the intentional inclusion of many loop candidates within multiple cities and reverse revisits with month-long time gaps. We chose two sequences: Riverside 03 and Sejong 01.

\textbf{NCLT\cite{ref_nclt}}: The NCLT Dataset was capture by a Velodyne HDL-32E LiDAR. This dataset was collected to facilitate research focusing on long-term autonomous operation in changing environments. The sessions repeatedly explore the campus, both indoors and outdoors, on varying trajectories, and at different times of the day across different seasons. We chose two sequences: 20120820 and 20130405. The details and route directions of these sequences are summarized in Table I.%, and we use every five bins files as a point cloud for fast retrieval. 

If a ground truth pose distance between the query and the matched node is less than 5m, the detection is considered as a true positive. In total 50 adjacent nodes are excluded from the search. All experiments are performed on the same laptop with an Intel i7-10870H CPU at 2.20GHz and 16GB memory.
% The experiments for NDD are conducted with $K$ candidates from the KD-tree. 
\begin{table*}[!t]
\caption{The Details of Selected Datasets Lists\label{tab:table1}}
\centering
\begin{tabular}{|c||c||c||c|}
\hline
Dataset   & Sequence    & \begin{tabular}[c]{@{}l@{}}Path length (km) (\# revisits / \# total)\end{tabular} & Route Dir. on revisit \\ \hline
         {KTTTI}  & 00  & 3.71 (852 / 4541) & Same      \\ \cline{2-4} 
                  & 02  & 4.26 (309 / 4661)  & Both      \\ \cline{2-4} 
                  & 05  & 2.23 (493 / 2761)  & Same      \\ \cline{2-4}
                  & 06  & 1.63 (153 / 1101)  & Same      \\ \cline{2-4}
                  & 08  & 3.21 (332 / 4072) & Reverse   \\ \hline
        {MulRan} & Riverside 03 & 6.81 (6052 /10425)    & Reverse \\ \cline{2-4}
                  & Sejong 01    & 23.16 (8633 / 28728)   & Same   \\ \hline
        {NCLT}   & 20120820    & 6.01 (526 / 4001)        & Both   \\ \cline{2-4} 
                  & 20130405    & 4.53 (675 / 5059)        & Both   \\ \hline

\end{tabular}
\end{table*}
% \begin{list}{}{}
% \item{\url{http://www.latex-community.org/}} 
% \item{\url{https://tex.stackexchange.com/} }
% \item{\url{https://tex.stackexchange.com/} }
% \end{list}
\subsection{Competing Methods and Evaluation}
We employ the raw point clouds as the input to verify the performance of competing methods. Since NDD is a global descriptor, the performance of our approach is compared to other global descriptors using a 3D point cloud. We compare with Scan Context\footnote{https://github.com/irapkaist/scancontext} (SC)\cite{ref_sc}, Intensity Scan Context\footnote{https://github.com/wh200720041/iscloam} (ISC)\cite{ref_isc}, LiDAR Iris\footnote{https://github.com/JoestarK/LiDAR-Iris}\cite{ref_iris}, and M2DP\footnote{https://github.com/LiHeUA/M2DP}\cite{ref_m2dp}.

We use the default parameters of the available codes for SC, ISC, M2DP and IRIS. Specifically, we set parameters of SC as a 60 (sectors)$\times$ 20 (rings) descriptor, and maximum sensing range of LiDAR is 80m. The maximum sensing range of ISC ($60\times20$) is 50m.  M2DP sets $l = 8$ (number of circles), $t = 16$ (bins in a ring), $p = 4$ (the azimuth) and $q = 16$ (the elevation). IRIS sets 80 (radial direction) $\times$ 360 (angular direction) bins. The proposed NDD descriptor needs to set the number of $N_s$ (the azimuth) and $N_r$ (the radial angle), and $K$ for the number of candidates from the KD-tree. In our experiments, we optimize the parameters on KITTI 08 sequence and ﬁx the values as $N_r$ = 20,  $N_s$ = 60, $K=25$, and we set the loop closure threshold of NDD similarity as 0.65 for all tests. The ﬁnal descriptor is a $(2\times 20)\times 60$ matrix\footnote{The code of our method: https://github.com/zhouruihao1001/NDD}.  

The precision-recall curve,  F1 score and Extended Precision (EP) are used in our paper to evaluate the performance of different competitors. The F1 score and EP\cite{ref_ep} are deﬁned as:
\begin{equation}
\label{deqn_ex13}
\begin{array}{c}
F_{1}=2 \times (P \times R)/(P+R), \\

E P=0.5 \times \left(R_{P 100}+P_{R 0}\right),
\end{array}
\end{equation}
where $P_{R 0}$ is the precision at minimum recall, and $R_{P 100}$ is the max recall at 100\% precision. %EP is speciﬁcally designed metrics for place recognition algorithms\cite{ref_ep}.

\subsection{Experimental Results}
The performance of the descriptors is analyzed using the precision-recall curve as in Fig. 5. It is noted that our approach can describe a point cloud by using just a single feature (probability density score) or two features (probability density score and entropy). In some tests, we use probability density score only as a descriptor, and name this method as 1-scale NDD. Additionally, we also use the F1 score and Extended Precision (EP) shown in Tab. II to analyze the performance on KITTI datasets.
\begin{figure*}[!t]
\centering
\subfloat[KITTI 00]{\includegraphics[width=2.3in]{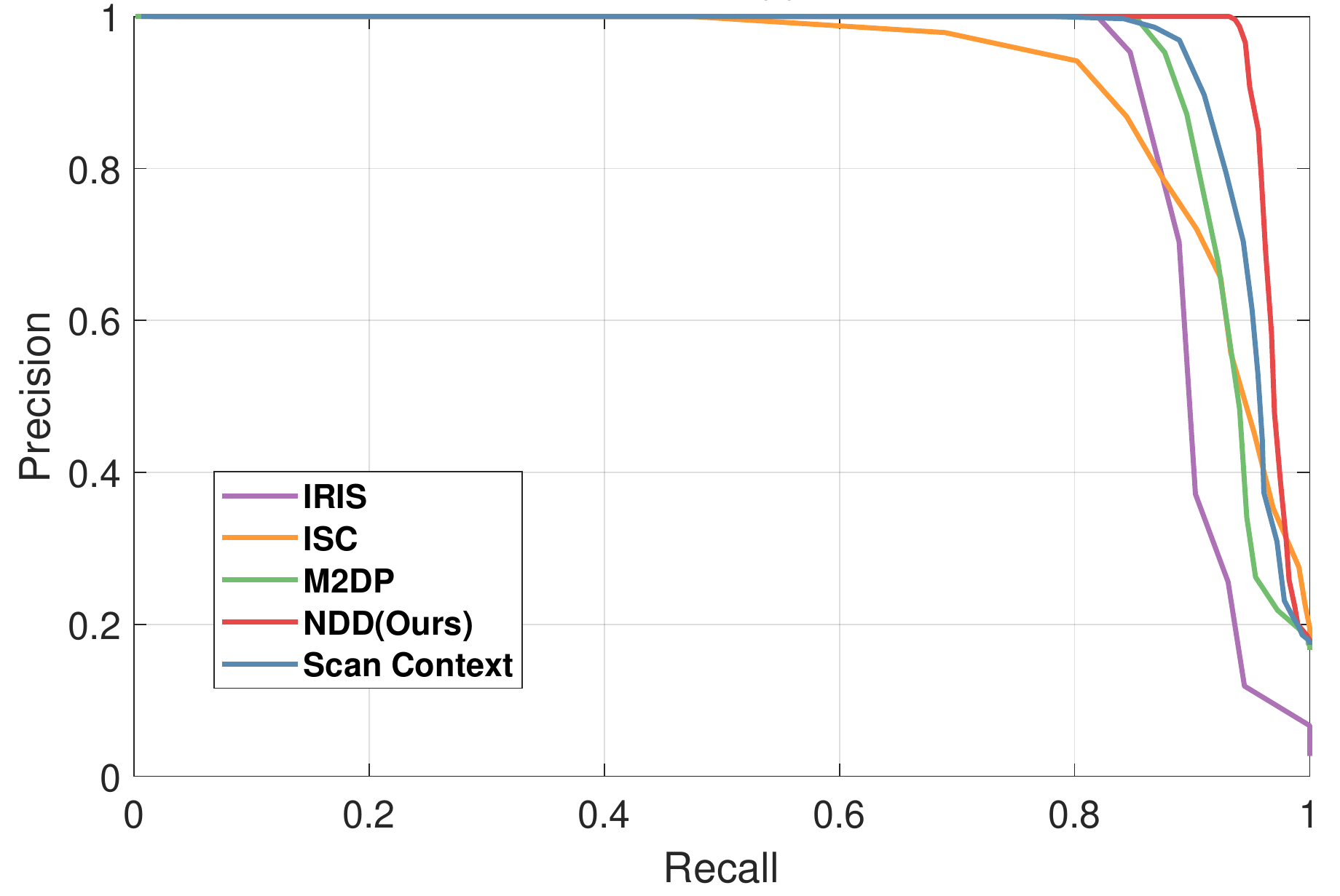}%
\label{fig_1_case}}
\subfloat[KITTI 02]{\includegraphics[width=2.3in]{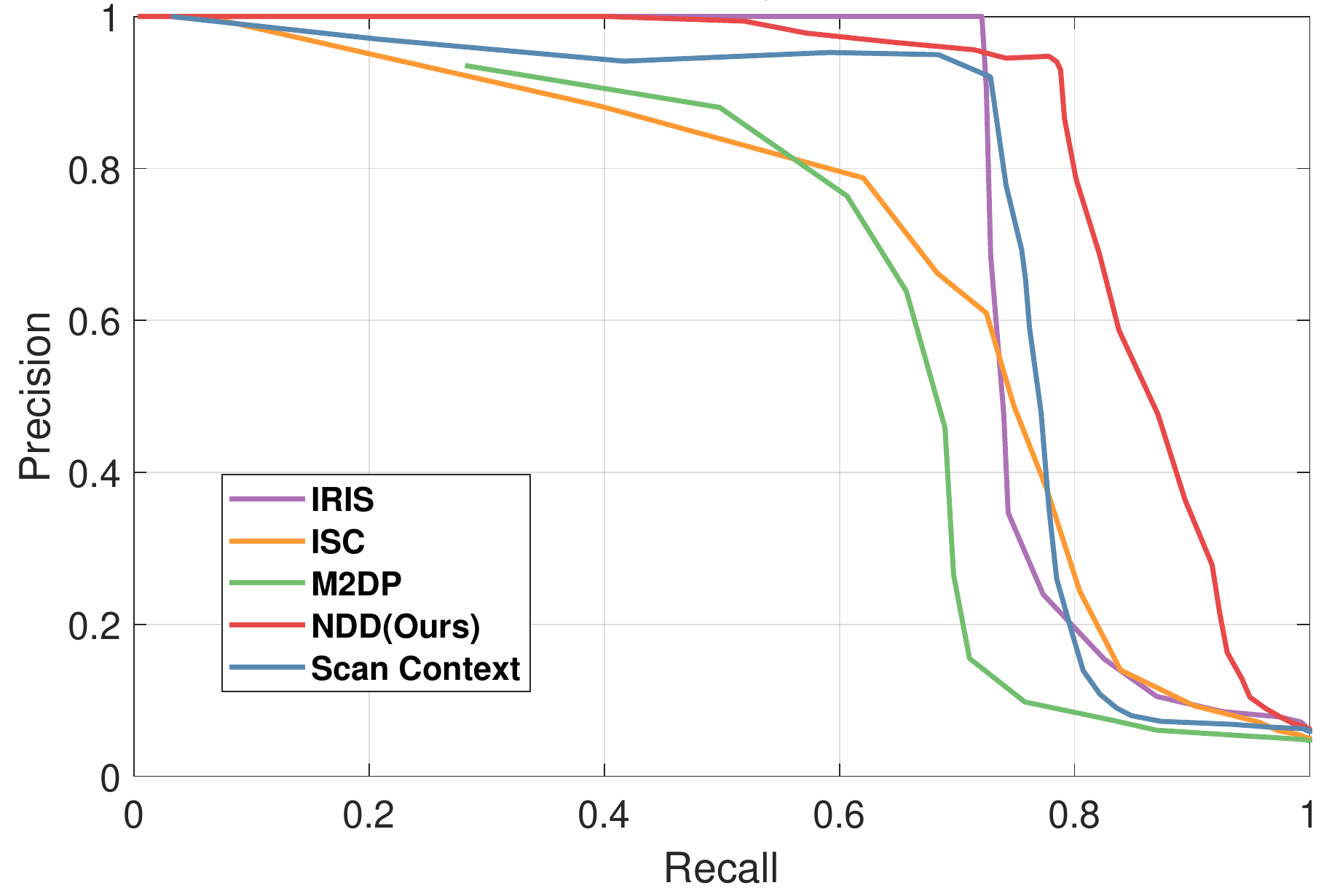}%
\label{fig_2_case}}
\subfloat[KITTI 05]{\includegraphics[width=2.3in]{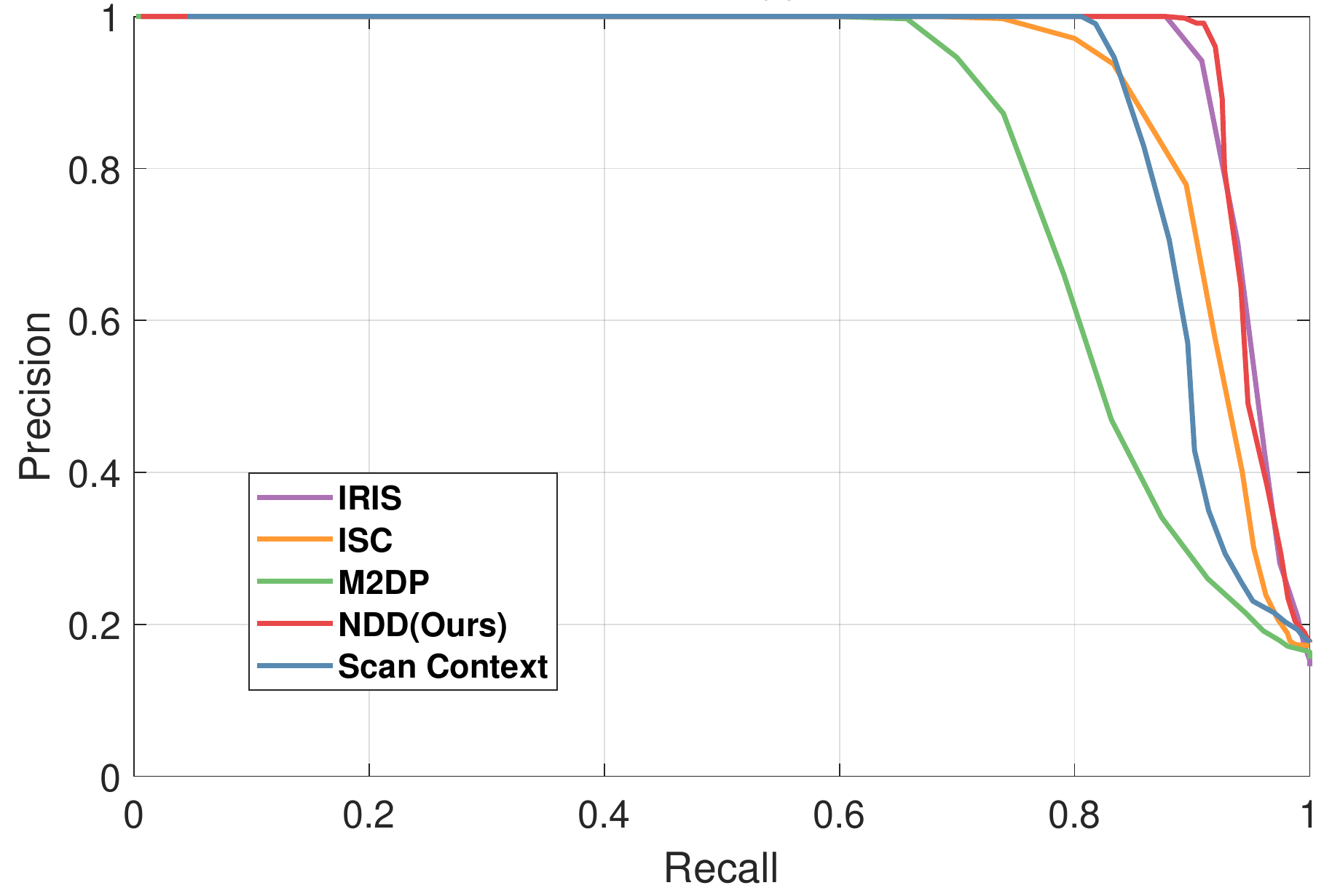}%
\label{fig_3_case}}
\hfil
\subfloat[KITTI 06]{\includegraphics[width=2.3in]{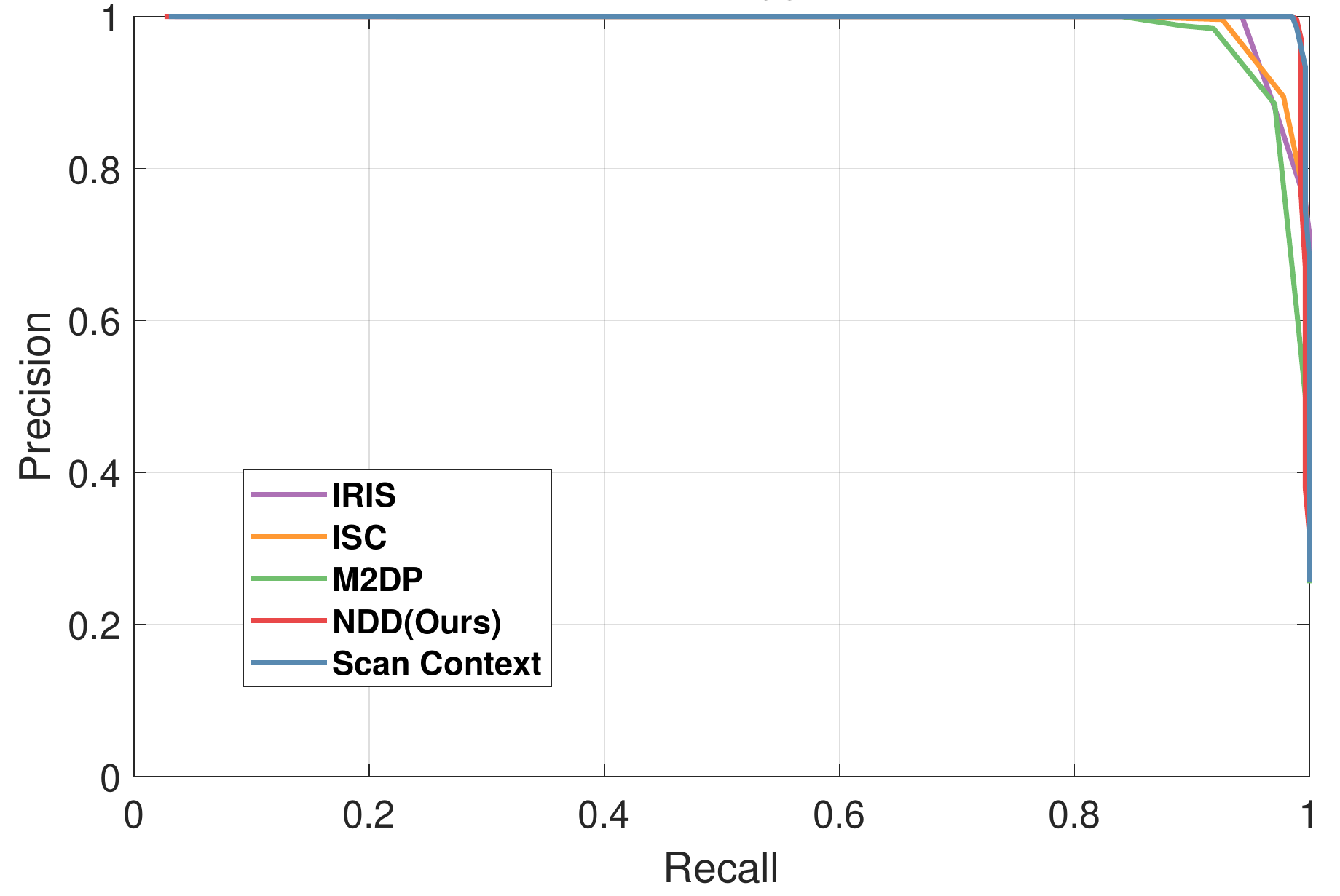}%
\label{fig_4_case}}
\subfloat[KITTI 08]{\includegraphics[width=2.3in]{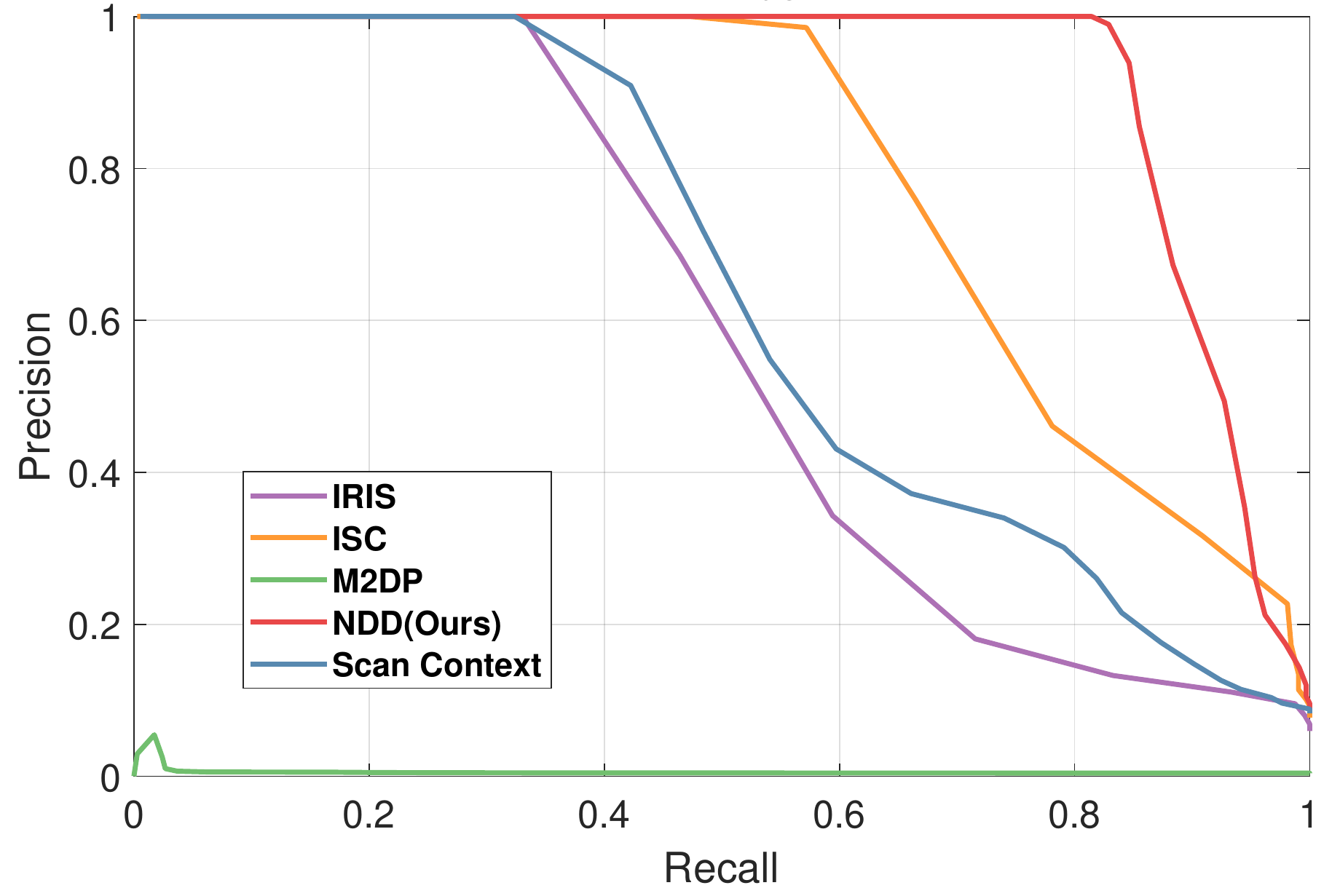}%
\label{fig_5_case}}
\subfloat[Sejong 01]{\includegraphics[width=2.3in]{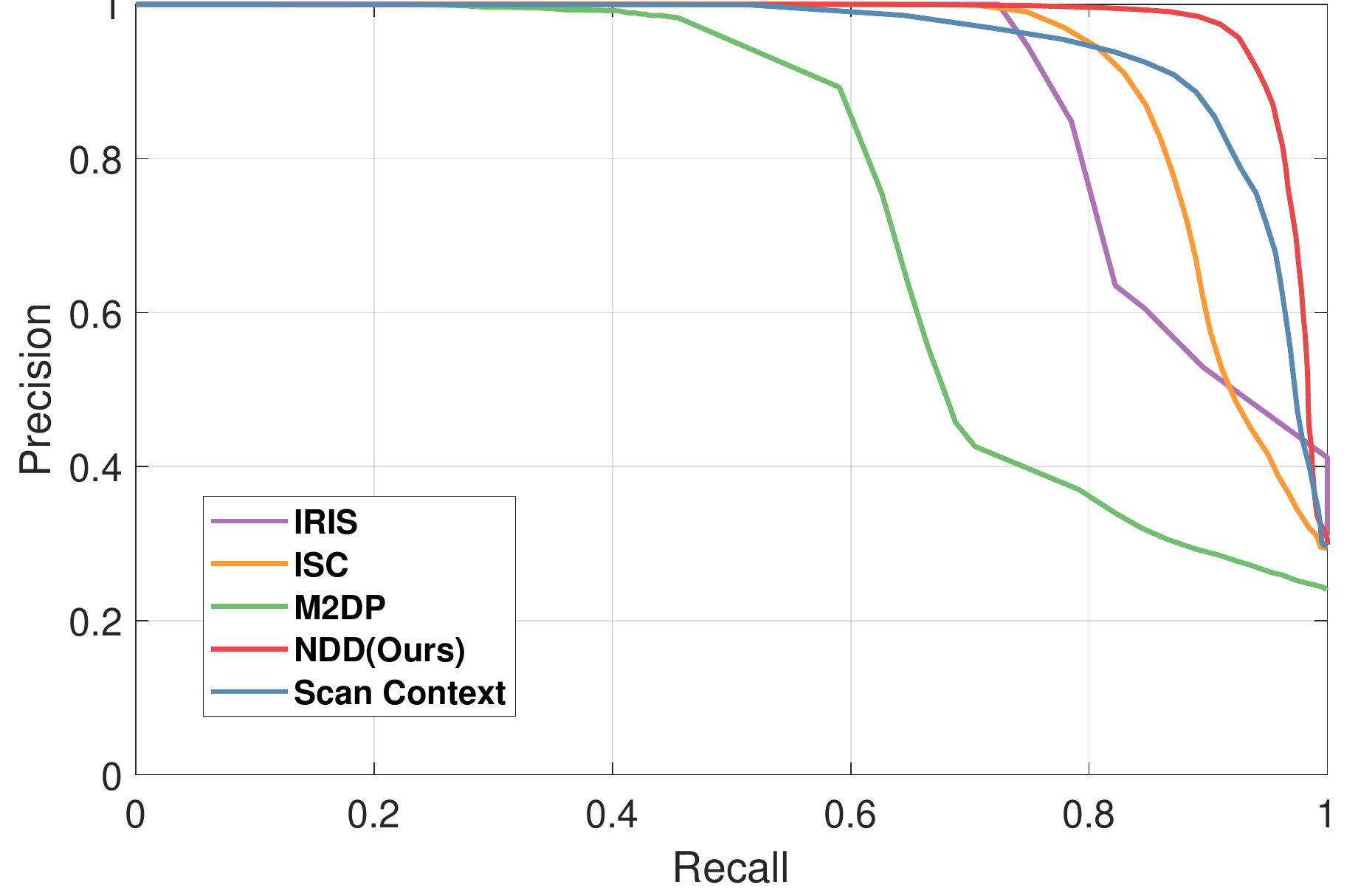}%
\label{fig_6_case}}
\hfil
\subfloat[Riverside 03]{\includegraphics[width=2.3in]{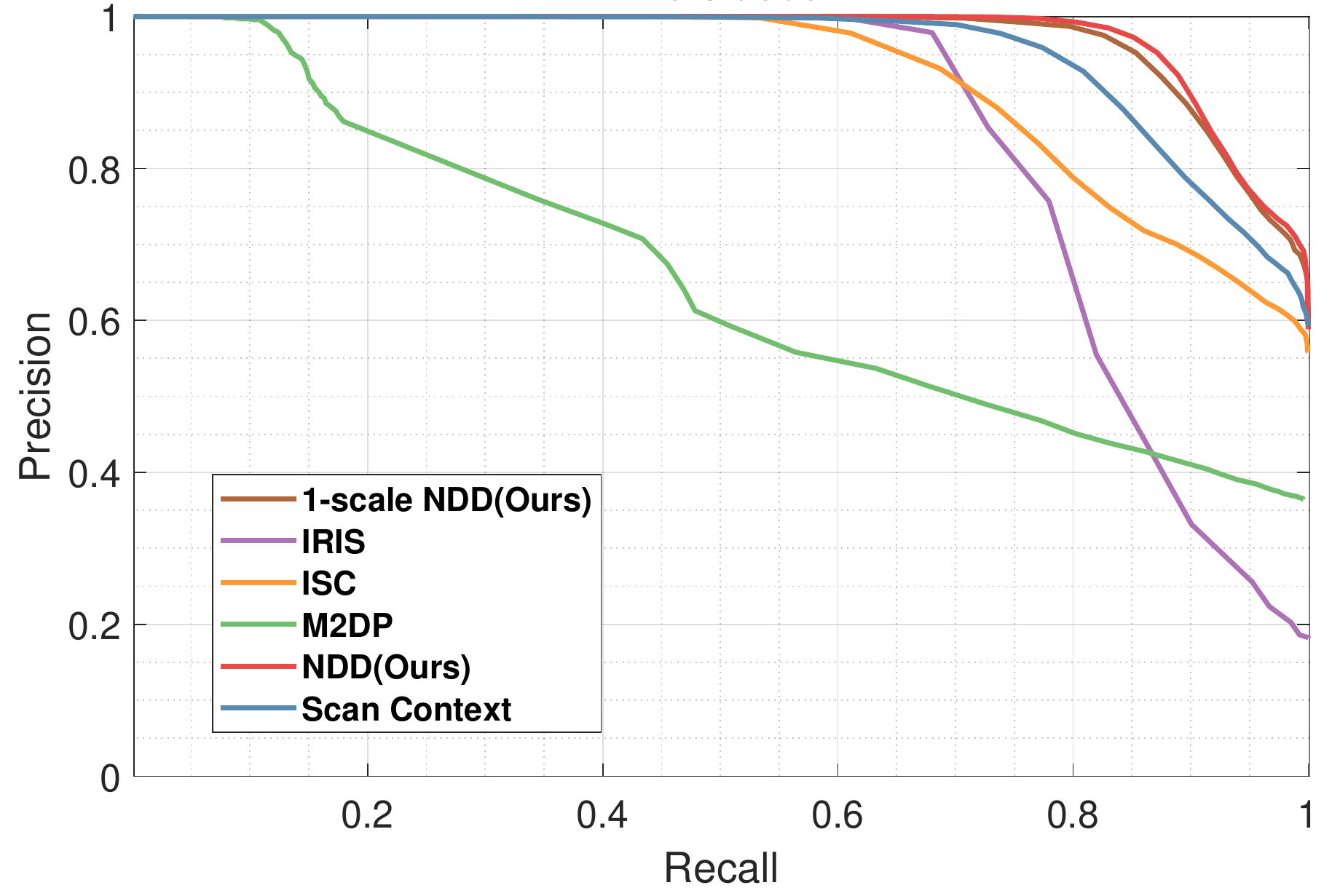}%
\label{fig_7_case}}
\subfloat[NCLT 20120820]{\includegraphics[width=2.3in]{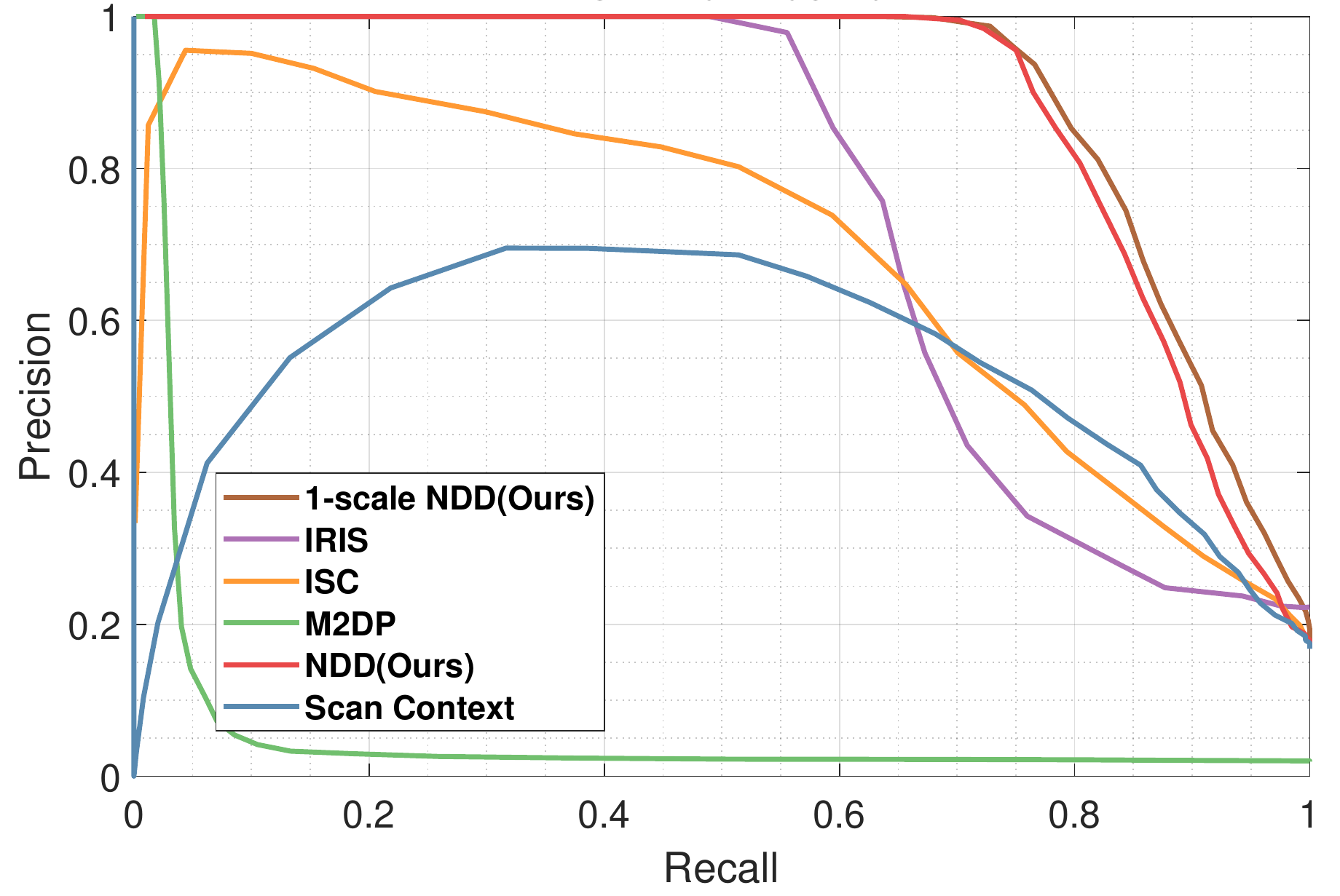}%
\label{fig_8_case}}
\subfloat[NCLT 20130405]{\includegraphics[width=2.3in]{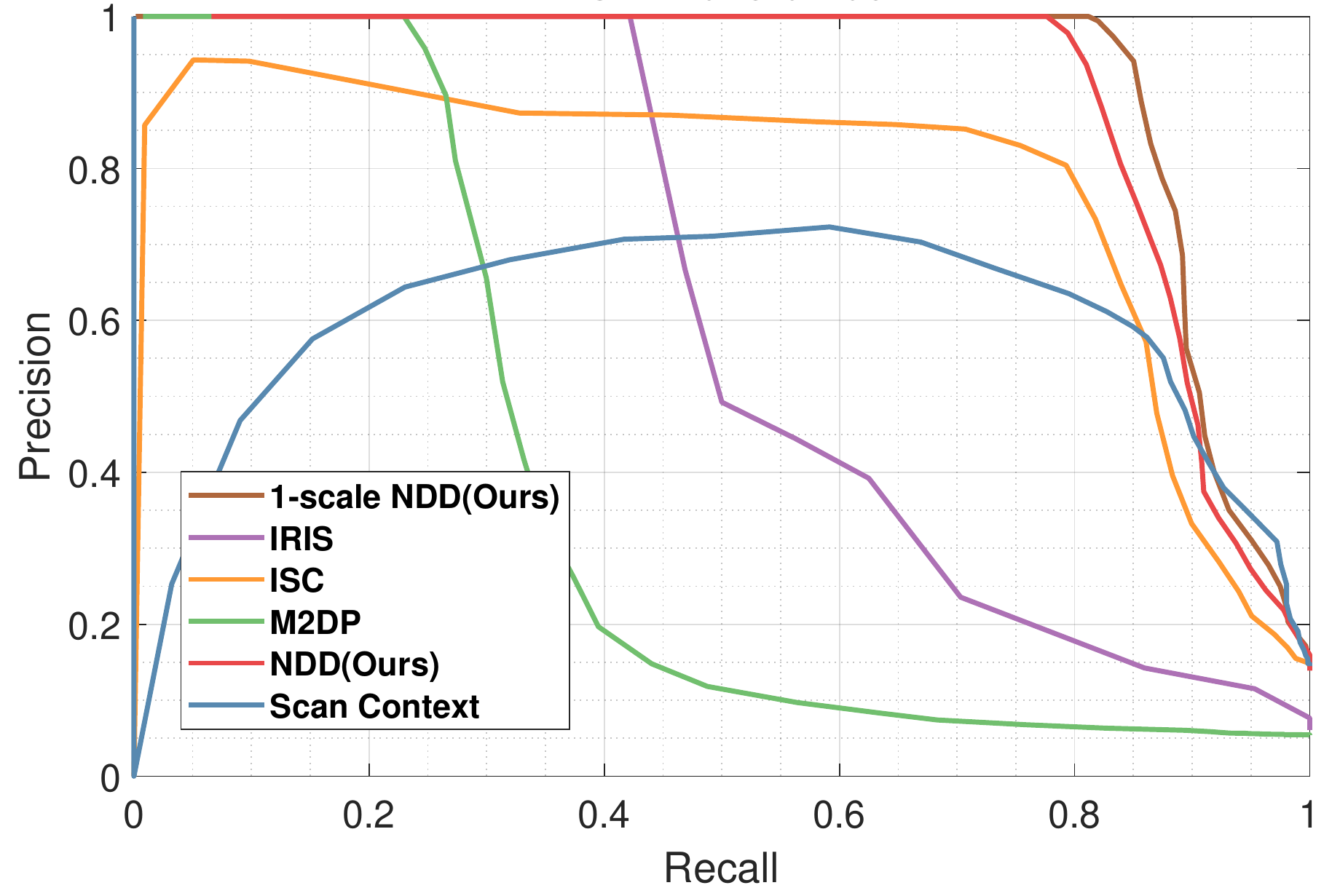}%
\label{fig_9_case}}

\caption{P-R curves on different dataset.}
\label{fig.5}
\end{figure*}

\begin{table*}[!t]
\caption{F1 score / Extended Precision on KITTI dataset\label{tab:table2}}
\centering
\begin{tabular}{|c||c||c||c||c||c||c|}
\hline
METHOD      & 00     & 02       & 05     & 06    & 07    & 08          \\ \hline
SC {[}9{]}   & 0.924/0.891 & 0.690/0.516 & 0.859/0.902 & 0.932/0.982 & 0.482/0.630 & 0.608/0.667 \\ \hline
ISC {[}10{]}           & 0.856/0.737 & 0.675/0.510 & 0.847/0.813 & 0.937/0.921 & 0.506/0.634 & 0.719/0.710 \\ \hline
IRIS {[}12{]}          & 0.873/0.909 & 0.813/\textbf{0.860} & 0.922/0.925 & 0.936/0.971 & 0.585/0.710 & 0.534/0.665 \\ \hline
M2DP {[}13{]} & 0.885/0.911 & 0.616/0.500 & 0.802/0.799 & 0.945/0.920 & 0.515/0.589 & 0.022/0.500 
\\ \hline
PointNetVLAD {[}14{]} & 0.779/0.641 & 0.727/0.691 & 0.541/0.536 & 0.852/0.767 & 0.531/0.591 & 0.037/0.500 \\ \hline
NDD (Ours)            & \textbf{0.943/0.963} & \textbf{0.846}/0.710 & \textbf{0.945/0.934} & \textbf{0.996/0.998} & \textbf{0.644/0.733} & \textbf{0.896/0.904} \\ \hline
\end{tabular}
\end{table*}

KITTI 00, KITTI 05 and KITTI 06 contain loops with limited rotations or lane changes. All methods on these three datasets have good performance for loop closure detection, as shown in Fig. 5. However, in the case of large rotation of point clouds, such as in KITTI 08 with a reverse loop, many methods, including M2DP, fail to detect the loops. One segment of KITTI 07 is incorrectly considered as closed loops in the ground truth due to the traffic jam for a while, so the performance of all descriptors is poor. Since one segment of the NCLT sequences contains a narrow indoor environment with little structural information useful, many methods do not work well in this case. 

Compared with our NDD, Scan Context is one-feature with the height value and shows good performance when applied to the outdoor urban datasets. However, the performance of SC is limited where vertical height change is trivial in non-urban environments (indoor environments), such as KITTI 08 and NCLT. Intensity Scan Context is also a one-feature encoding method with intensity. ISC performs better than SC in indoor environments. LiDAR Iris performs excellently in various outdoor environments, but it requires high retrieval time cost during matching process, as shown in TABLE IV. The proposed approach (NDD) significantly outperforms other descriptors, especially in reverse loop detection. This illustrates the proposed method is robust to rotation even for a reverse revisit.

\textbf{SLAM Integration.}
One benefit of our method is CPU-friendly and can be implemented in real time. We integrate our NDD into existing LiDAR odometry and mapping\cite{ref_loam} (LOAM) system for loop closure detection, as shown in Fig. 6. NDD-LOAM is accurate for loop closure detection and corrects the cumulative error of trajectory during mapping.

\begin{figure}[!t]
\centering
\subfloat[Lego-LOAM]{\includegraphics[width=3.2in]{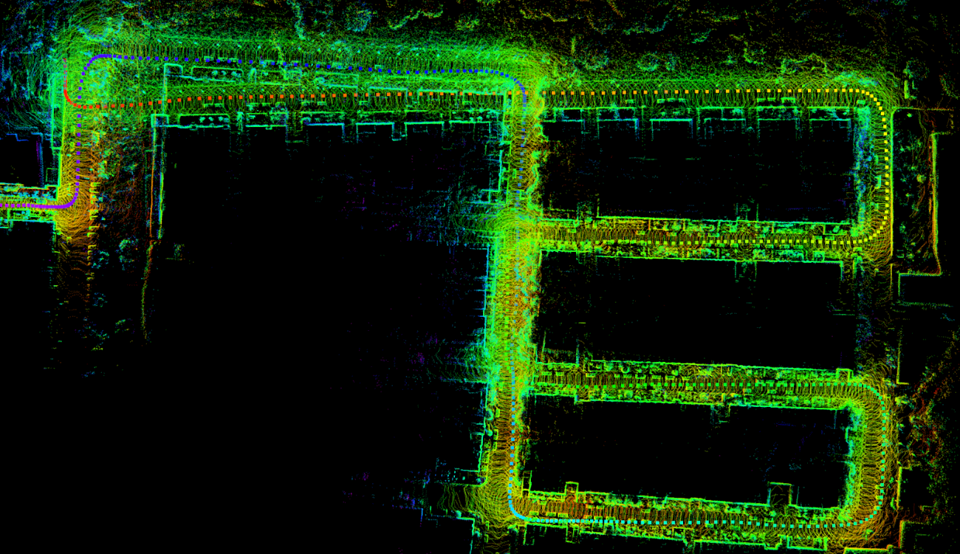}%
\label{fig_6_a}}
\hfil
\subfloat[NDD-LOAM]{\includegraphics[width=3.2in]{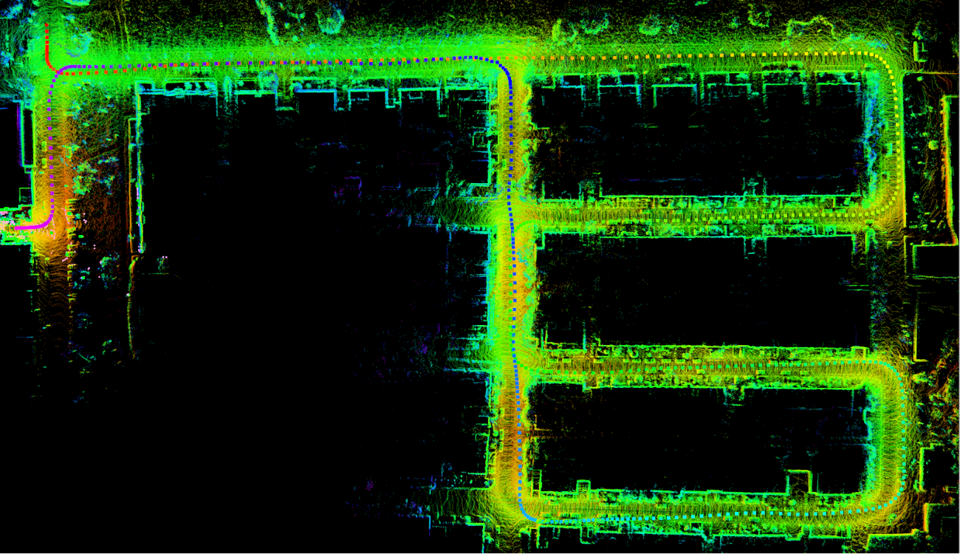}%
\label{fig_6_b}}

\caption{SLAM integration. (a) Lego-LOAM  and (b) NDD-LOAM results on KITTI 08 with the proposed NDD LiDAR loop closure detection.}
\label{fig.8}
\end{figure}

\subsection{Ablation Studies}
\textbf{Effects of Encoding Function}: We validate the effectiveness of different encoding strategy by comparing with $H$ (maximum height, used in SC), $P$ (probability density score) and $E$ (entropy) as mentioned in Sec. III. B. 
% Scan Context\cite{ref_sc} shows both good performance and efficiency by H-encoding, 
We replace $H$ with $P$ and $E$ on KITTI 08 dataset to verify the performance of $P$, $E$, $H$ and the proposed $P+E$. The results are shown in TABLE III in which $P+E$ means our method NDD and others are one-feature descriptor. $P$-encoding shows better performance than $H$-encoding on KITTI 08. A combination of $P$ and $E$ outperforms any one-feature methods in this test, indicating the superiority of our NDD over others.

\textbf{Effects of Matching Method}: We adopt the SC-fashion cosine similarity (cos) and our correlation coefficient similarity (corr) in this test. The effectiveness of the matching method is verified by comparing the average cosine distance (cos) and correlation coefficient (corr), as shown in Tab. III. We also show the PR curve of different matching and encoding methods in Fig. 7. On average, the proposed correlation coefficient similarity is 8.2\% higher than that of cosine similarity in terms of F1, and 5.4\% higher in terms of EP.

\begin{table}[!t]
\caption{Ablation Studies on KITTI 08 (F1 score / EP) \label{tab:table3}}
\centering
\resizebox{\linewidth}{!}{
\begin{tabular}{|c||c||c||c||c|} 
\hline
Factor   &$H$          &$E$         &$P$      &$P+E$ \\
            % & (1-scale)  &(1-scale)  &(1-scale)   &(2-scale) \\
\hline
cos         &0.608/0.667		&0.621/0.678  &0.729/0.765	&0.860/0.869  \\
\hline
corr         &0.701/0.720   &0.724/0.711    &0.826/0.859    &0.896/0.904  \\ 
\hline
\end{tabular} 
}
\end{table}

\begin{figure}[!t]
\centering
\includegraphics[width=3.2in]{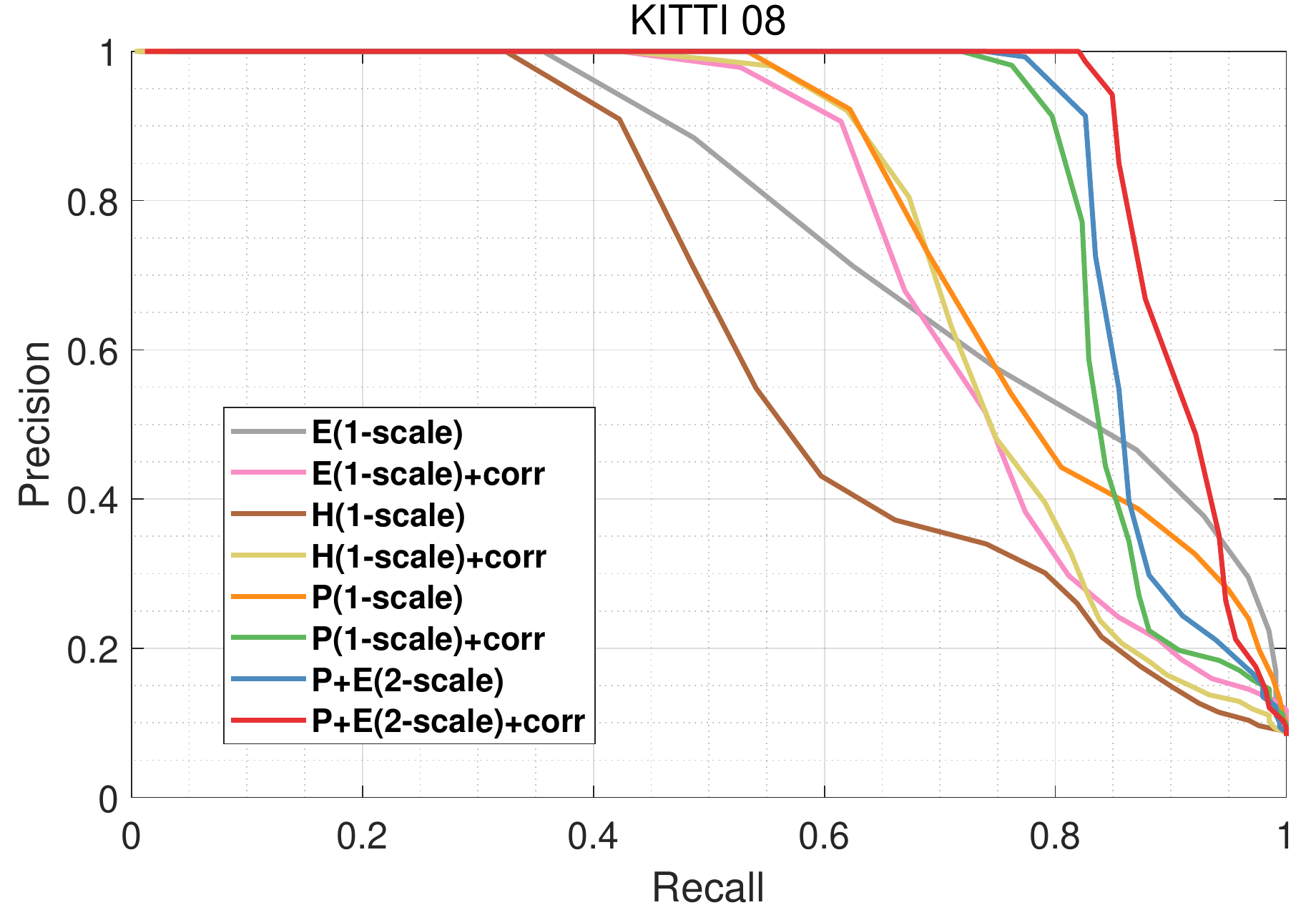}
\caption{The performance of different matching methods and encoding functions. ($H$ is maximum height, $P$ is probability density score, $E$ is entropy, and corr means correlation coefficient. $P+E$ means the two-scale NDD and others are one-scale descriptor)}
\label{fig_7}
\end{figure}

\subsection{Time Cost}
\textbf{Overall Time Cost}: We compare our method with other methods in terms of overall time cost on the KITTI datasets. The time cost of one method consists of two parts: descriptor generation (description) and loop closure detection (retrieval). The average time cost on KITTI 08 is given in Table IV. As shown in Tab. IV, IRIS requires the highest time cost. This is due to the fact that IRIS retrieval compares all candidate key frames. M2DP is also time-consuming in this test since the eigen and SVD decomposition have high complexity. SC requires higher time cost than that of ISC and ours. ISC and the proposed NDD take similar time cost but our accuracy, as shown in Tab. II, is much higher than ISC.

\begin{table*}[!t]
\caption{Average Time Cost ON KITTI 08\label{tab:table6}}
\centering
%  \resizebox{\linewidth}{!}{
\begin{tabular}{|c||c||c||c||c|}
\hline
Methods 	&Size	&Description (ms)	&Retrieval (ms) &Total (ms)\\
\hline
SC [9] 	&20 × 60	&48	&55	&103\\
\hline
ISC [10] 	&20 × 60	&50 &26 &76\\
\hline
IRIS [12]	&360 × 80	&31  &5502  &5533\\
\hline
M2DP [13] 	&192 × 1	&175  &5 &180\\
\hline
1-scale NDD (Ours) 	&20 × 60	&60	&8	&68\\
\hline
2-scale NDD (Ours) 	&20 × 60 × 2	&62	&10 &72\\
\hline
\end{tabular} 
% }
\end{table*}

\textbf{Time Cost of Rotation Alignment}: In this section, we compare three rotation alignment methods on KITTI 00 dataset. The competing alignment methods are:
\begin{enumerate}

\item{Original column-shifting: we compare directly two full descriptors by column-shifting.}

\item{Binarize: we binarize two full descriptors $D_A$ and $D_B$ and compute the Hamming distance between two matrices after binarization. To binarize a full descriptor $D$, all elements of non-zero values in $D$ are set to 1 and the remaining are 0. Then we calculate similarity between two binary matrices by XNOR operation. Finally, we find the maximum similarity by column-shifting the binary matrices.
}

\item{Row Vector (Ours): we covert a full descriptor to a $1 \times N_{s}$ row vector as mentioned in Sec. IV. B. 
}
\end{enumerate}

TABLE V shows that the average time costs of different methods. The first two methods essentially compare the similarity of matrices, while our method compares the similarity of two vectors. Compared with the first two methods, our method requires less time cost of alignment.
\begin{table}[!t]
\caption{Average Time Cost of Different Strategy\label{tab:table4}}
\centering
\begin{tabular}{|c||c||c||c|}
\hline
Method	&Origin	&Binary	&Row Vector\\
\hline
Retrieval Time Cost (ms)	&51.22	&22.34	&9.78  \\
\hline

\end{tabular}
\end{table}

\textbf{Time Cost of Retrieval}: We test the influence of different retrieval strategies (searching key and full descriptor retrieval) on a large dataset, Mulran sejong 01. The F1 scores of searching key and full descriptor retrieval are close, which are respectively 0.846 and 0.850. However, the average time costs are 35ms and 1567ms, respectively on this dataset. The computational cost of the full descriptor retrieval is 45 times higher than that of the key searching method.

% \begin{table}[!t]
% \caption{Influence Of Retrieval Strategy\label{tab:table5}}
% \centering
% \begin{tabular}{|c||c||c|}
% \hline
% Method of Retrieval	&Time cost(s)	&F1 score\\
% \hline
% Full descriptor	&43887	&0.850\\
% \hline
% Key	&1003	&0.846\\
% \hline
% \end{tabular}
% \end{table}

\section{Conclusion}
In this paper, we present a descriptor based on normal distribution for a 3D point cloud, and apply it to LiDAR-based loop closure detection. As a global descriptor, the proposed approach employs the probability density score and entropy to describe the local shape of one point cloud. We also propose a fast retrieval method by both key searching and fast rotation alignment. After retrieval, we use correlation coefficient to calculate the similarity between two descriptors. Experimental results show that the proposed NDD method performs better than several STOA methods in terms of both accuracy and efficiency.

% \newpage

 

% \end{IEEEbiography}



\vfill

\end{document}